\documentclass[11pt,a4paper]{article}
\usepackage[hyperref]{acl2021}
\usepackage{times}
\usepackage{latexsym}
\usepackage{xcolor}
\usepackage{graphicx}

\usepackage{microtype}
\usepackage{hyperref}
\usepackage{soul}

\aclfinalcopy 

\title{IITK@LCP at SemEval 2021 Task 1:  Classification for Lexical Complexity Regression Task}

\author{
    Neil Rajiv Shirude$^{*}$ \qquad   
    Sagnik Mukherjee\thanks{\quad Authors equally contributed  to this work.} \\
    \large{\textbf{Tushar Shandhilya}} \qquad
    \large{\textbf{Ananta Mukherjee}} \qquad
  \large{\textbf{Ashutosh Modi}} \\
{Indian Institute of Technology Kanpur (IIT Kanpur)} \\
{\tt \{neilrs, sagnikm, anantam, stushar\}@iitk.ac.in}\\
  {\tt ashutoshm@cse.iitk.ac.in}  \\
}
\date{}

\begin{document}
\maketitle
\begin{abstract}
This paper describes our contribution to SemEval 2021 Task 1:  Lexical Complexity Prediction. In our approach, we leverage the ELECTRA model and attempt to mirror the data annotation scheme. Although the task is a regression task, we show that we can treat it as an aggregation of several classification and regression models. This somewhat counter-intuitive approach achieved an MAE score of 0.0654 for Sub-Task 1 and MAE of 0.0811 on Sub-Task 2. Additionally, we used the concept of weak supervision signals from Gloss-BERT in our work, and it significantly improved the MAE score in Sub-Task 1. 

\end{abstract}

\section{Introduction}




With the rapid growth in digital pedagogy, English has become an extremely popular language. Although English is considered an easy language to learn and grasp, a person's choice of words often affects texts' readability. The use of difficult words can potentially lead to a communication gap, thus hampering language efficiency. Keeping these issues in mind, many Natural Language Processing tasks for text simplification have been recently proposed \cite{10.5555/3207692.3207704,sikka2020survey}. Our task of lexical complexity prediction is an important step in the process of simplifying texts.\\
The SemEval 2021 Task 1 \cite{shardlow2021semeval} focuses on lexical complexity prediction in English. Given a sentence and a token from it, we have to predict the complexity score of the token. The task has two Sub-Tasks-\\
\textbf{Sub-Task 1:} complexity prediction of single words \\
\textbf{Sub-Task 2:} complexity prediction of multi word expressions (MWEs). \\
A word might seem complex because of 2 major factors-\\  \textbf{a)} The word is less common or complex in itself. \\ \textbf{b)} The context in which the word is used makes it hard to comprehend. \\
Observing the orthogonality of these two reasons, we captured the context-dependent features and independent features separately, trained models on them individually, and then combined the two using ensemble methods. We used the ELECTRA \cite{clark2020electra} model for extracting context-dependent features and GloVe embeddings \cite{pennington2014glove} for representing the word-level features. \\
Additionally, we propose a classification pipeline that is trained on GloVe embeddings of the tokens. This pipeline can be interpreted as a model for capturing different annotators' thought processes: overconfidence, under-confidence and randomness. We are making our code available for our models and experiments via GitHub\footnote{\href{https://github.com/neilrs123/Lexical-Complexity-Prediction}{  https://github.com/neilrs123/Lexical-Complexity-Prediction}}.


\section{Background}

This task uses the CompLex dataset \cite{shardlow-etal-2020-complex}, which is a lexical complexity prediction dataset in English for single and multi word expressions (2-grams). Sentences in this task consists of sentences taken from 3 corpora- Bible, Biomed and Europarl. The train, validation and test split of the data was  9179, 520, 1103 respectively. We used the trial data as the validation set. \\
The aim of the task is to predict how complex a given token in a given sentence is. More mathematically, given a tuple $[s,t,c]$, where $s = [t_1, t_2, ...t_n]$ and $t = t_{j}$, we have to give an estimate of the function $\sigma$, such that $\sigma(s,t) = c$. (s is the sentence, t is the token and c is the complexity score).\\
The earlier focus on this task has been through the SemEval 2016 Task 11 (\citet{paetzold-specia-2016-semeval}). However, it was a binary classification task. Most of the participating systems used Support Vector Machines such as \citet{kuru-2016-ai} and \citet{choubey-pateria-2016-garuda}, decision trees and random forests (\citet{choubey-pateria-2016-garuda}, \citet{brooke-etal-2016-melbourne}, \citet{ronzano-etal-2016-taln}), and even basic threshold based approaches (\citet{kauchak-2016-pomona}, \citet{malmasi-etal-2016-ltg}). Very few of them, including \citet{bingel-etal-2016-coastalcph} used neural networks. The system by \citet{wrobel-2016-plujagh} achieved an F1 score very close to the winning solution using only single feature - word frequency from Wikipedia. Most of these systems use word embeddings, POS information and word frequencies as features. The winning system by \citet{paetzold-specia-2016-sv000gg} however uses 69 morphological, semantic and syntactic features.\\
Another related shared task was presented at the BEA workshop at 2018 \cite{yimam-etal-2018-report}. It had a probabilistic task as well as a binary classification task. Even there, the organizers conclude that feature engineering has worked better than neural networks. The winning system by \citet{gooding-kochmar-2018-camb} uses feature engineering and later random forest and linear regression models.

\section{System Overview}
Our proposed pipeline can be divided into the following 4 main components-\\ \textbf{a)} Feature Extraction \\\textbf{b)} Regression Pipeline \\ \textbf{c)} Classification Pipeline \\\textbf{d)} Ensemble\\
The pipeline is shown in Figure \ref{fig:x}.

\begin{figure}
    \centering
    \includegraphics[height=4cm ,width=8 cm]{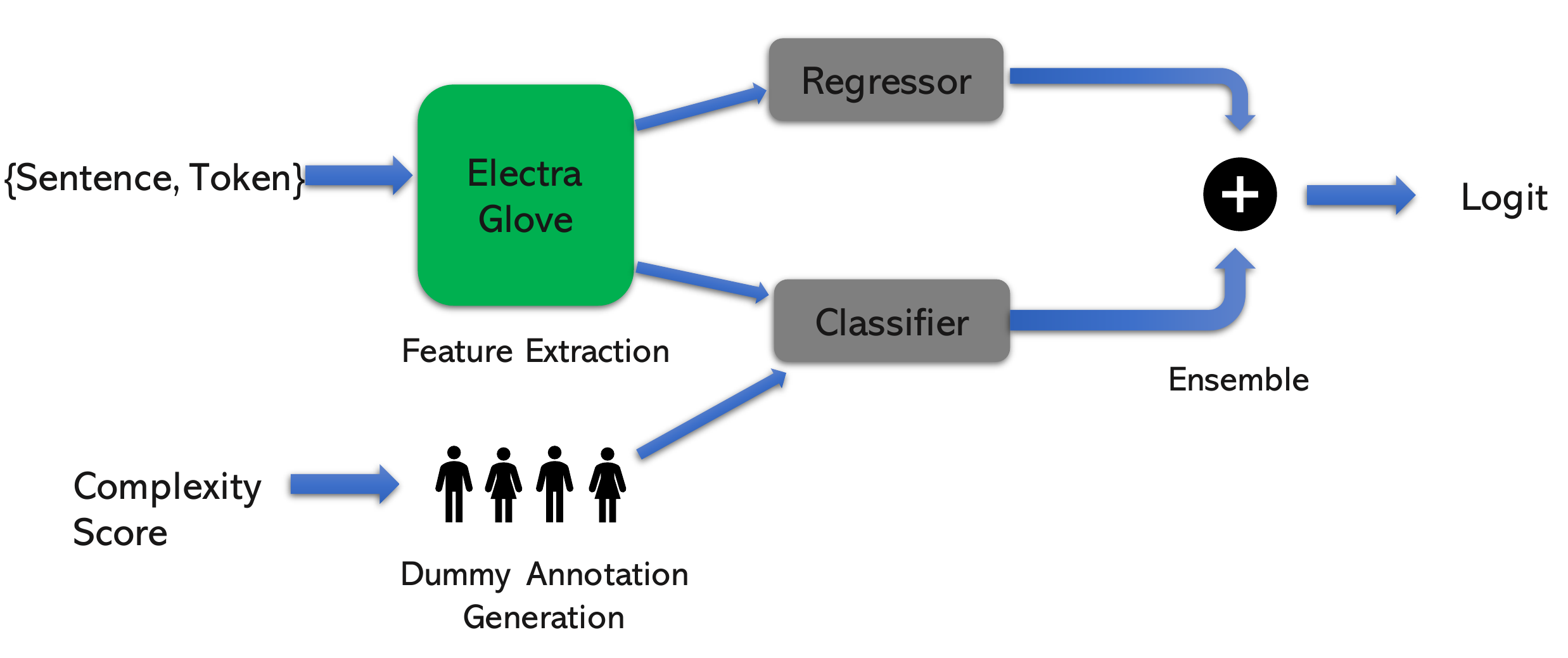}
    \caption{Solution Pipeline} 
    \label{fig:x}
\end{figure}

\subsection{Feature Extraction}

ELECTRA is a transformer based model, that is trained like a discriminator and not like generator. And in our case, this model performed exceptionally well on the validation data as compared to BERT \cite{devlin2019bert}.

We extracted context-dependent features using embeddings generated from the ELECTRA model and captured context-independent word-level features using static 200-dimensional GloVe embeddings of the tokens. \\
In order to generate the embeddings of the target word through ELECTRA, we implemented the KMP pattern matching algorithm \cite{enwiki:1005503556} to find the indices of the sub-tokens of the target token in the tokenized sentence. Subsequently, we calculated an average across these sub-token embeddings generated by ELECTRA.\\
While using GloVe embeddings, in the case of multi-word expressions in Sub-Task 2, the average of the embeddings of both token words was taken as the feature vector. If a word was not present in the GloVe dictionary, the GloVe embedding was initialized to a 200-dimensional vector consisting of zeros. 

\begin{figure}
    \centering
    \includegraphics[height=4cm ,width=7 cm]{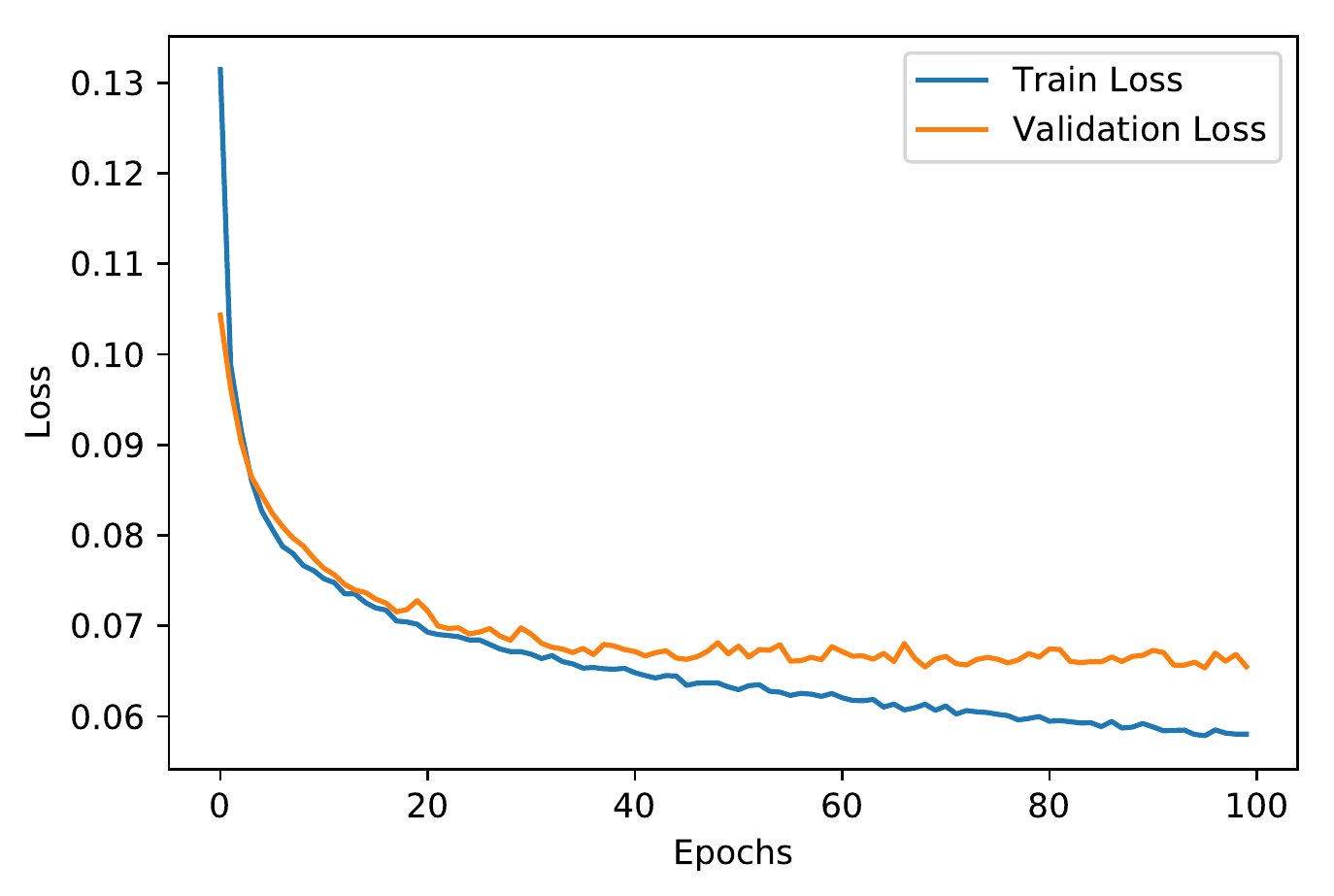}
    \caption{Convergence of losses for finetuning ELECTRA with weak supervision}
    \label{fig:x}
\end{figure}

\subsection{Regression Pipeline}
The most natural way to look at the lexical complexity prediction task is to treat it as a regression task. The regression pipeline, a significant component of our system, is based on this idea. For Sub-task 1, in the regression pipeline, a pretrained ELECTRA model was finetuned with a linear layer on top of it. We leveraged the model directly available at the Huggingface library \cite{wolf2020huggingfaces}. Only the last transformer layer of ELECTRA was kept trainable. The remaining ones were kept frozen. For Sub-task 2, a fixed ELECTRA model (non-trainable weights) was used to generate token embeddings and a linear regression model was trained with these extracted embeddings.

\noindent\textbf{Weak Supervision: } 
In order to have higher attention on the target word, the use of weak supervision signals proved useful. Inspired by GlossBert \cite{huang-etal-2019-glossbert}, the target word was wrapped with single inverted commas (' 's) as a weak signal to the transformer \cite{vaswani2017attention} model. This technique significantly improved the results obtained using the regression pipeline in subtask I. However, the same technique applied to subtask II made the scores worse.  

\begin{table}[h!]
\centering
\begin{tabular}{|p{.1\textwidth}|p{.15\textwidth}|p{.15\textwidth}|}
\hline
Method & Val MAE & Test MAE \\[1 ex]
\hline\hline
+ signal & 0.06516 & 0.06800 \\
\hline
- signal & 0.06990 & 0.07118 \\
\hline
\end{tabular}
\caption{Variation of MAE scores with and without the signalling technique for Sub-task 1: the single word task. ('+ signal' means weak supervision has been used and '- signal' means otherwise.)}
\label{table:final}
\end{table}

\subsection{Classification Pipeline}
\noindent\textbf{Motivation from Annotation Procedure: }
Another way to look at the task is via a novel classification pipeline that is inspired from the data annotation process that is explained in \citet{shardlow-etal-2020-complex}. Even though the task is a regression task, each data annotator performed a 5 class classification-\\
Given a sentence and a token in the sentence, each annotator had to select one class from among Very Easy, Easy, Neutral, Difficult and Very Difficult. Each of these classes was mapped to a discrete label between 0 and 1- namely 0, 0.25, 0.5, 0.75 and 1 respectively. The final complexity score was an average of up to 20 such annotations.\\
The Classification Pipeline aims to model this data annotation procedure. The main idea of this process is to teach classification models how to annotate data tuples. The three main components of this scheme are-\\
\textbf{a)} Generating dummy annotations from complexity scores \textbf{b)} Training classification models on dummy annotations, and \textbf{c)} Aggregating all predicted annotations to generate predicted complexity scores.

\noindent\textbf{Generation of Dummy Annotations: }
A given complexity score can be represented as a weighted average of its lower and upper target classes and the weights can be determined using the magnitude of the complexity score. These weights then determine the proportions of the two classes in the set of dummy annotations for that data tuple. For example, if the number of dummy annotators is $n=5$ and the complexity score of the training example is $c=0.2$, the lower and upper target classes are $low=0$ and $high=0.25$, respectively. Let $\alpha$ be the proportion of dummy annotations with the lower target class. Correspondingly, $1-\alpha$ will be the proportion with the upper target class. The number of dummy annotations with $target\_class = low$ are given as $floor(n*\alpha)$ and that with $target\_class = high$ as $n-floor(n*\alpha)$. $\alpha$ can be calculated using the equation-
\begin{center}
    $c = \alpha*low + (1-\alpha)*high$
\end{center}
We get $\alpha=0.2$. Hence, we have $floor(n*\alpha) = 1$ dummy annotations with $target\_class = low (0)$ and remaining 4 annotations with $target\_class = high (0.25)$. Hence, the dummy annotations set for $c=0.2$ is ${0,0.25,0.25,0.25,0.25}$. Similarly, the dummy annotations set for $c=0.8$ is ${0.75,0.75,0.75,0.75,1}$.\\
In this process, we also attempted to capture the impact of intentional human errors made during the data annotation procedure. Just like a weary or uninterested annotator who would have randomly selected for one of the five classes for a certain data tuple, a small fraction of the dummy annotations was assigned random values from the set containing 0, 0.25, 0.5, 0.75 and 1. This modification aims to model the small-scale randomness in annotation procedure. \\
Using this procedure, dummy annotation sets of size $n$ can be generated for any value of $c$, where $n$ can be treated as a hyperparameter. The value $n$ can also be interpreted as the number of classification models that are being trained in the next step. 



\noindent\textbf{Classification Models: }
In a diverse set of annotators, there will be over-confident annotators who will select lower classes and there will be under-confident annotators who will select upper classes. Then there will be neutral annotators as well. By ensuring that the dummy annotations are sorted, we can say that the first classifier learns how to annotate like the over-confident annotator, the last classifier learns how to annotate like the under-confident annotator and the classifiers in between model the neutral annotators. We trained SVM classifiers with RBF kernels, using GloVe embeddings of token words as features.

\noindent\textbf{Aggregation of Predicted Annotations:} The annotations were aggregated by simply taking the average of all predicted class labels in order to obtain the final predicted complexity scores. Each of these models may have high individual variances, but the ensemble tends to have lower variance and bias. Also, any number of models can be inserted in the ensemble without leading to over-fitting on the train data.

\subsection{Ensemble}
In order to have a better bias variance trade off and also to exploit the ``expertise" of different pipelines, the final approach incorporates both the regression and classification pipelines to form an ensemble. The final predicted complexity was obtained by taking an ensemble of the predictions from the regression and classification pipelines as described above. The classification pipeline for both the Sub-Tasks was based on GloVe embeddings as features and SVM classifiers. The regression pipeline for Sub-Task 1 was based on fine-tuning ELECTRA with weak supervision and that for Sub-Task 2 was based on features collected from ELECTRA model (non-trainable) with a linear regression trained on it.

\section{Experimental Setup}
\begin{figure}
    \centering
    \includegraphics[height=4cm ,width=7.5 cm]{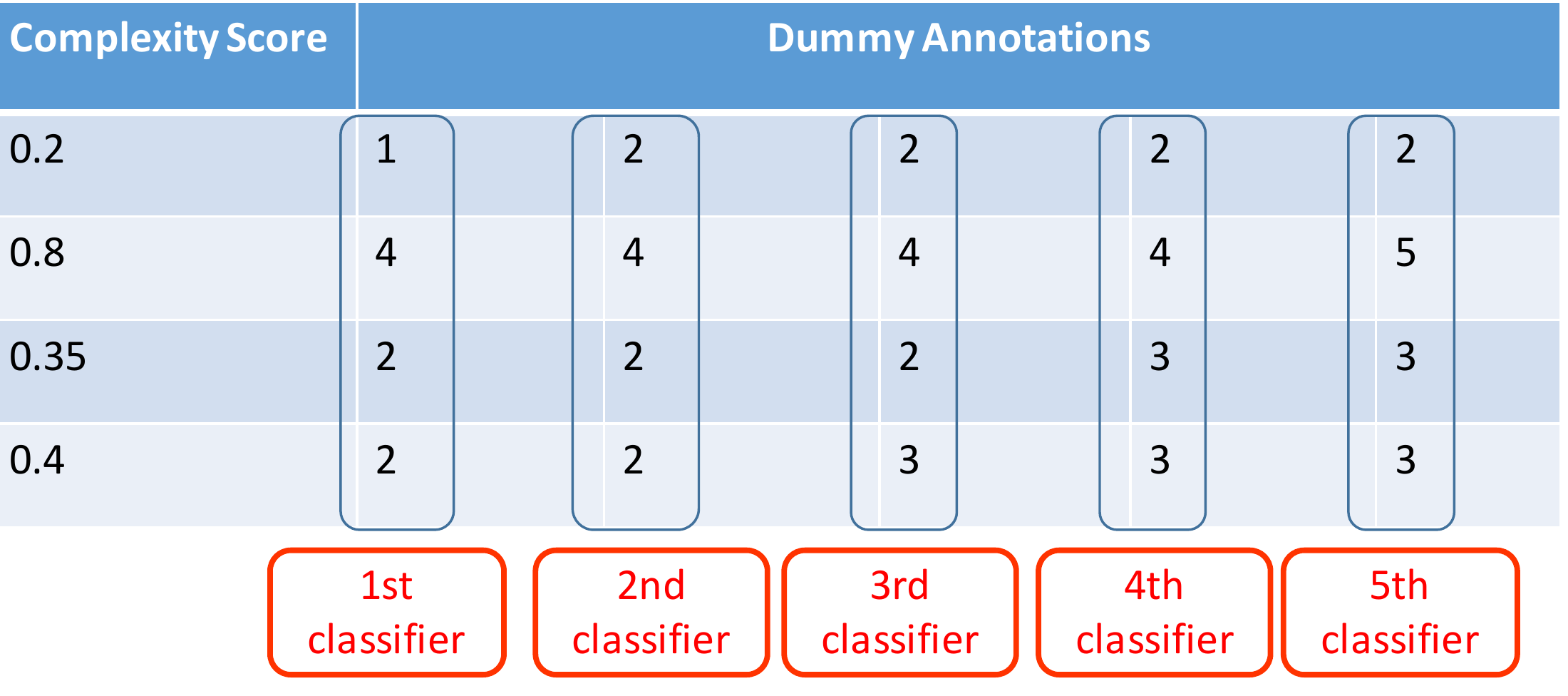}
    \caption{A few worked out examples of generating dummy annotations from complexity scores. For each of these cases, the continuous labels 0,0.25,0.50,0.75 and 1 are mapped to categorical labels 1,2,3,4,5 and then put into SVM. Clearly the labels of the 1st classifier is less that that of the second one. i.e. on a scale of confidence, the first classifier is at a lesser position. So it models a less confident person.} 
    \label{fig:x}
\end{figure}

The official evaluation metric for both the Sub-Tasks was Pearson Correlation (standard for regression tasks). For both sub-tasks, the train/test/val split as per the official release has been used. The ELECTRA finetuning was done with an NVIDIA GTX 1080 GPU with early stopping (93 epochs). We used the MAE loss function to train the model with an adam optimizer with $lr = 1e^-5$, $eps=1e-08$ and $weight decay=0$ . Training set was shuffled and the batch size was kept at 64. In the ELECTRA model, the padding parameter was set to True and maximum length was at 140. For the SVM models the value of slack was chosen to be  1 and for SVM and Linear regresion the sklearn \cite{JMLR:v12:pedregosa11a} library was used. All the hyperparameters were tuned with a grid search method.
\section{Results}



\noindent\textbf{Results on Validation Data:} The comparison of the baseline results and our results obtained using the regression pipeline, the classification pipeline and the ensemble of the two models on the validation set (trial data) is given in Table \ref{table:reg}.  

\begin{table}[h!]
\centering
\begin{tabular}{|p{.063\textwidth}|p{.07\textwidth}|p{.07\textwidth}|p{.07\textwidth}|}
\hline
Task & Baseline & Regre-& Classi-\\
& & ssion & fication\\
& & Pipeline & Pipeline\\
\hline\hline
Subtask 1 & 0.0853 & 0.0651 & 0.0641 \\
\hline
Subtask 2 & - & 0.0840 & 0.0768  \\
\hline
\end{tabular}
\caption{Results on validation set (Mean Absolute Errors)}
\label{table:reg}
\end{table}

\begin{table}[h!]
\centering
\begin{tabular}{|p{.063\textwidth}|p{.07\textwidth}|p{.07\textwidth}|p{.07\textwidth}|p{.082\textwidth}|}
\hline
Task & MAE & Pearson & MSE \\

\hline\hline
One & 0.0623 & 0.8308 & 0.0065 \\
\hline
Two & 0.0727 & 0.8146 & 0.0087  \\
\hline
\end{tabular}
\caption{Results on Vaditation Set for final ensemble}
\label{table:reg}
\end{table}

\noindent\textbf{Results on Test Data:} Our results on the test data along with the best results obtained for each task are shown in Table \ref{table:final}.\\
The winning system's pearson and MAE scores on the test data are as follows: 0.7886 and 0.0609 for subtask I(single word expressions), 0.8612 and 0.0616 for subtask II(multi word expressions).

\begin{table}[h!]
\centering
\begin{tabular}{|p{.063\textwidth}|p{.07\textwidth}|p{.07\textwidth}|p{.07\textwidth}|p{.082\textwidth}|}
\hline
Task & MAE & Pearson & MSE \\

\hline\hline
One & 0.0654 & 0.7511 & 0.0071 \\
\hline
Two & 0.0811 & 0.8277 & 0.0098  \\
\hline
\end{tabular}
\caption{Results on Test Set}
\label{table:reg}
\end{table}



\section{Error Analysis}
\begin{figure}
    \centering
    \includegraphics[height=3.5cm ,width=7.5 cm]{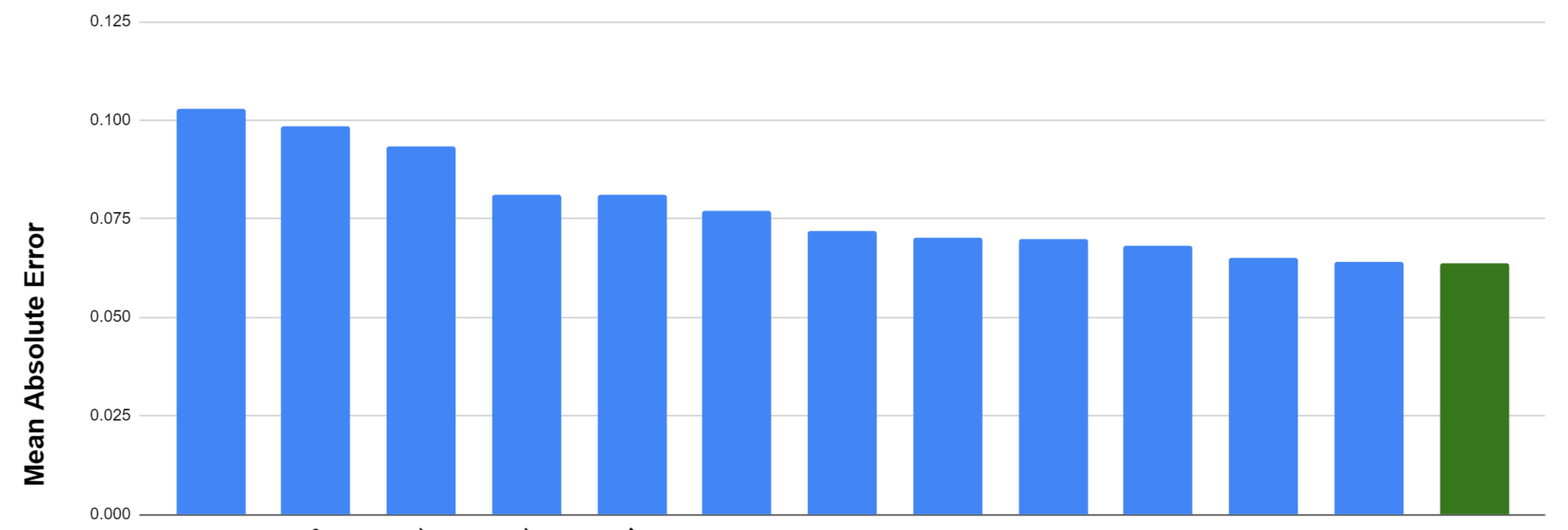}
    \caption{Comparison of MAE values of the models we tried (Subtask I). From left the models are (1) Linear Regression with hand crafted features, (2) Character level RNN, (3) Character level CNN, (4) and (5) sentence and character level GRU and LSTMs (6), (7), (8), (9) and (10) Linear Regression with GloVe, ELECTRA and BERT embeddings, (11) the current regression pipeline, (12) classification pipeline, (13) Final ensemble   } 
    \label{fig:5}
\end{figure}

Analyzing all the experiments and the corresponding results, the following can be concluded: 
\textbf{a)} Word-level features as well as context-dependent features need to be considered while determining complexity of a token. \textbf{b)} Approaches based on the data annotation scheme are well suited to tackle the lexical complexity prediction task. 
\textbf{c)} Ensemble of a large number of simple models is an effective way of tackling this task.  
\textbf{d)} Models with large number of parameters like BERT \cite{} suffer heavily due to overfitting, where as ELECTRA base prove to be much better.\\
The model architectures that were tried out in earlier stages showed similar trends. For example, ELECTRA finetuning produced much better scores than BERT finetuning. Also, simpler models like a simple linear regression on GloVe embeddings showed promise, proving that simpler models with lesser parameters worked better. All these trends across those models are visually shown in Figure  \ref{fig:5}. It was observed that the model was underperforming on the tuples from Biomed corpus. However the scores did not improve using BERT variants like BioBERT \cite{Lee_2019}, BioMedBERT \cite{chakraborty-etal-2020-biomedbert} and a few other transformer based models pretrained on biomedical texts. A variant of ELECTRA on biomedical texts could have improve on this, however due to its unavailability it could not be tried out. \\
In majority of the prior work on LCP, there is abundance use of word frequency as a feature. However, in this system the scores got worse when frequency features were used along with others in ensemble. And the feature in itself could not produce competitive results. Previously, \citet{gong2020frage} and \citet{mu2018allbutthetop} have shown that frequency information causes significant distortion in the embedding space. We also hypothesize that the frequency information in GloVe embeddings help us in this regard.

\section{Conclusion}
In this paper we presented a system for lexical complexity prediction in the form of a regression task. The proposed system's primary novelty is in  treating it as a classification task and trying to model the annotation scheme. An ensemble of these classification models and vanilla fine-tuning of ELECTRA model proved to be very useful. Also the weak supervision based approach gave the scores a significant boost for the Sub-Task 1. 

\bibliographystyle{acl_natbib}
\bibliography{anthology,acl2021}


\end{document}